\documentclass{article}

\PassOptionsToPackage{numbers, compress}{natbib}
\usepackage[final,nonatbib]{nips_2017} 

\usepackage[utf8]{inputenc} 
\usepackage[T1]{fontenc}    
\usepackage{hyperref}       
\usepackage{url}            
\usepackage{booktabs}       
\usepackage{amsfonts}       
\usepackage{nicefrac}       
\usepackage{microtype}      
\usepackage{subfigure}
\usepackage[labelfont=bf,font=normalsize]{caption}

\usepackage{amsmath}
\usepackage{amsfonts}
\usepackage{amssymb}
\usepackage{enumerate}
\usepackage{float}
\usepackage{bm}
\usepackage{graphicx}
\usepackage{natbib}
\usepackage{multirow}
\usepackage{dirtytalk}
\usepackage{array}

\usepackage{algorithm}
\usepackage{algorithmic}

\newcommand{\comment}[1]{}

\DeclareMathOperator*{\argmin}{arg\,min}
\DeclareMathOperator*{\expectation}{\mathbb{E}}

\title{Annealed Generative Adversarial Networks}

\author{Arash Mehrjou \\
Department of Empirical Inference \\
Max Planck Institute for Intelligent Systems\\
\texttt{arash.mehrjou@tuebingen.mpg.de} \\
\And
Bernhard Sch\"{o}lkopf \\
Department of Empirical Inference \\
Max Planck Institute for Intelligent Systems\\
\texttt{bs@tuebingen.mpg.de} \\
\And
Saeed Saremi \\
Redwood Center for Theoretical Neuroscience\\
University of California, Berkeley \\
\texttt{saeed@berkeley.edu} \\
}

\begin{document}

\maketitle


\begin{abstract}
We introduce a novel framework for adversarial training where the target distribution is annealed between the uniform distribution and the data distribution. We posited a conjecture that learning under continuous annealing in the nonparametric regime is stable irrespective of the divergence measures in the objective function and proposed an algorithm, dubbed $\beta$-GAN, in corollary. In this framework, the fact that the initial support of the generative network is the whole ambient space combined with annealing are key to balancing the minimax game. In our experiments on synthetic data, MNIST, and CelebA, $\beta$-GAN with a fixed annealing schedule was stable and did not suffer from mode collapse. 
\end{abstract}

\section{Introduction} \label{sec:INTRO}
{\it Background---} One of the most fundamental problems in machine learning is the unsupervised learning of high-dimensional data. A class of problems in unsupervised learning is density estimation, where it is assumed that there exist a class of probabilistic models underlying observed data $x$ and the goal of learning is to infer the \say{right} model(s). The generative adversarial network proposed by Goodfellow {\it et al.}~\citep{goodfellow2014generative} is an elegant framework, which transforms the problem of density estimation to an adversarial process in a minimax game between a generative network $\mathbb{G}$ and a discriminative network $\mathbb{D}$. However, despite their simplicity, GANs are notoriously difficult to train.

{\it Mode collapse---} There are different schools in diagnosing and addressing the problems with training GANs, that have resulted in a variety of algorithms, network architectures, training procedures, and novel objective functions~\citep{radford2015unsupervised,salimans2016improved,zhao2016energy, arjovsky2017wasserstein, nowozin2016f}. The roots of the problems in training GANs lie on the unbalanced nature of the game being played, the difficulty with high-dimensional minimax optimizations, and the fact that the data manifold is highly structured in the ambient space $\mathcal{X}$. Perhaps, the biggest challenge is that the natural data in the world reside on a very low-dimensional manifold of their ambient space~\citep{narayanan2010sample}. Early in training the generative network $\mathbb{G}$ is far off from this low-dimensional manifold and the discriminative network $\mathbb{D}$ learns quickly to reject the generated samples, causing little room to improve $\mathbb{G}$. This was analyzed in depth by Arjovsky \& Bottou~\citep{arjovsky2017towards}, which highlighted the deficiencies of $f$-divergences when the generative network has a low-dimensional support. The other challenging issue is that GANs' optimal point is a {\it saddle point}. We have good understanding and a variety of optimization methods to find local minima/maxima of objective functions, but {\it minimax} optimization in high-dimensional spaces have proven to be challenging.  Because of these two obstacles, i.e. the nature of high-dimensional data and the nature of the optimization, GANs suffer from stability issues and the ubiquitous problem of {\it mode collapse}, where the generator completely ignores parts of the low-dimensional data manifold.

{\it $\beta$-GAN---} In this work, we address these two issues at the same time by lifting the minimax game, where the initial objective is to find the GAN equilibrium in an \say{easier} game of learning to map $z\sim p(z)$ to $x_0 \sim {\rm Uniform}[-1,1]^d$. Here, $z$ is the noise variable corresponding to the latent space, and $d$ is the dimension of the ambient space $\mathcal{X}$. The subscript in $x_0$ refers to the \say{inverse temperature} $\beta=0$, which is defined in the next section. After arriving at the equilibrium for $\beta=0$, we anneal the uniform distribution towards the data distribution while performing the adversarial training simultaneously. Our assumption in this work is that once GAN is stable for the uniform distribution, it will remain stable in the continuous annealing limit irrespective of the divergence measure being used in the objective function. In this work, we used the original Jensen-Shannon formulation of Goodfellow {\it et al.}~\citep{goodfellow2014generative}.  The objective to learn the uniform distribution puts constraints on the architecture of the generative network, most importantly $\dim(z) \geq d$, which has deep consequences for the adversarial training as discussed below.

{\it Related works---}  There are similarities between our approach here and recent proposals in stabilizing the GAN training by adding noise to samples from the generator and to the data points~\citep{kaae2016amortised,arjovsky2017towards}. This was called \say{instance noise} in~\citep{kaae2016amortised}. The key insight was provided in~\citep{arjovsky2017towards}, where the role of noise was to enlarge the support of the generative network and the data distribution, which leads to stronger learning signals for the generative network during training. The crucial difference in this work is that we approached this problem from the perspective of annealing distributions and our starting point is to generate the uniform distribution, which has the support of the {\it whole ambient space} $\mathcal{X}$. This simple starting point is a straightforward solution to theoretical problems raised in~\citep{arjovsky2017towards} in using $f$-divergences for adversarial training, where it was assumed that the support of the generative network has measure $0$ in the ambient space $\mathcal{X}$. Since the uniform distribution is not normalized in $\mathcal{R}^{d}$, we assumed $\mathcal{X}$ to be a finite $d$-dimensional box in $\mathcal{R}^{d}$. A good physical picture to have is to imagine the data manifold diffusing to the uniform distribution like ink in a $d$-dimensional vase filled with water.  \comment{To generate the uniform distribution in the GAN framework, the dimension $d$ of the latent space $z$ must be at least equal to the dimension of the ambient space $\mathcal{X}$ (See Lemma 1 in~\citep{arjovsky2017towards}). This is in contrast with current practices in adversarial training, where $\dim(z)$ is typically (much) less than $d$.}What $\beta$-GAN achieves during annealing is to shape the space-filling samples, step-by-step, to samples that lie on the low-dimensional manifold of the data distribution. Therefore, in our framework, there is no need to add any noise to samples from the generator (in contrast to~\citep{kaae2016amortised,arjovsky2017towards}) since the generator support is initialized to be the ambient space. \comment{To generate the uniform distribution in the GAN framework $\dim(z)\geq d$ (See Lemma 1 in~\citep{arjovsky2017towards}). We discuss the limitations and possible solutions for large datasets in the discussion section.} Finally, one can also motivate $\beta$-GAN from the perspective of curriculum learning~\citep{bengio2009curriculum}, where learning the uniform distribution is the initial task in the curriculum.
\section{$\beta$-GAN} \label{sec:MODEL}
In this section, we define the parameter $\beta$, which plays the role of inverse temperature and parametrizes annealing from the uniform distribution ($\beta=0$) to the data distribution ($\beta=\infty$). We provide a new algorithm for training GANs based on a conjecture with stability guarantees in the continuous annealing limit. We used the Jensen-Shannon formulation of GANs~\citep{goodfellow2014generative} below but the conjecture holds for other measures including $f$-divergences~\citep{nowozin2016f} and the Wasserstein metric~\citep{arjovsky2017wasserstein}. 

We assume the generative and discriminative networks $\mathbb{G}$ and $\mathbb{D}$ have very large capacity, parameterized by deep neural networks $G(z;\theta_G)$ and $D(x;\theta_D)$. Here, $z \sim p(z)$ is the (noise) input to the generative network $G(z;\theta_G)$, and $D(x;\theta_D)$ is the discriminative network that is performing logistic regression. The discriminative network is trained with the binary classification labels $D=1$ for the $N$ observations $\{x^{(1)}, x^{(2)}, \cdots, x^{(N)}\} \in \mathcal{R}^{d}$, and $D=0$ otherwise. The GAN objective is to find $\theta^*_G$ such that $G(z;\theta^*_G) \sim p_{\rm data}(x)$. This is achieved at the Nash equilibrium of the following minimax objective:
\begin{eqnarray}
\theta^*_G &=& \argmin_{\theta_G} \max_{\theta_D} f(\theta_D,\theta_G),\\
f(\theta_D,\theta_G) &=& \expectation_{x\sim p_{\rm data}} \log\left(D(x;\theta_D)\right) +  \expectation_{z\sim p(z)} \log(1-D(G(z;\theta_G);\theta_D)),
\end{eqnarray}
where at the equilibrium $D(G(z;\theta^*_G);\theta^*_D)=1/2$~\citep{goodfellow2014generative}. One way to introduce $\beta$ is to go back to the empirical distribution and rewrite it as a mixture of Gaussians with zero widths:
\begin{equation}
p_{\rm data}(x) = \frac{1}{N} \sum_i \delta(x-x^{(i)}) = \frac{1}{N}  \lim_{\beta \rightarrow \infty} \sqrt{\frac{\beta}{2\pi}} \sum_i \exp\left(- \frac{\beta (x-x^{(i)})^2}{2}\right).
\end{equation}
The heated data distribution at finite $\beta$ is therefore given by:
\begin{equation}
p_{{\rm data}}(x;\beta) =  \frac{1}{N} \left(\frac{\beta}{2\pi}\right)^{d/2} \sum_i \exp\left(- \frac{\beta (x-x^{(i)})^2}{2}\right).
\end{equation}

\newpage

{\it The $d$-dimensional box---} The starting point in $\beta$-GAN is to learn to sample from the uniform distribution. Since the uniform distribution is not normalized in $\mathcal{R}^{d}$, we set $\mathcal{X}$ to be the finite interval $[a,b]^d$.  The uniform distribution sets the scale in our framework, and the samples $x_\beta \sim p_{{\rm data}}(x;\beta)$ are rescaled to the same interval. This hard $d$-dimensional \say{box} for the data \say{particles} is thus assumed throughout the paper. Its presence is conceptually equivalent to a diffusion process of the data particles in the box $[a,b]^d$, where they diffuse to the uniform distribution like ink dropped in water~\citep{sohl2015deep}. In this work, we parametrized the distributions with $\beta$ instead of the diffusion time. We also mention a non-Gaussian path to the uniform distribution in the discussion section.  

\begin{figure}[t!]
\centering
\includegraphics[width=9cm]{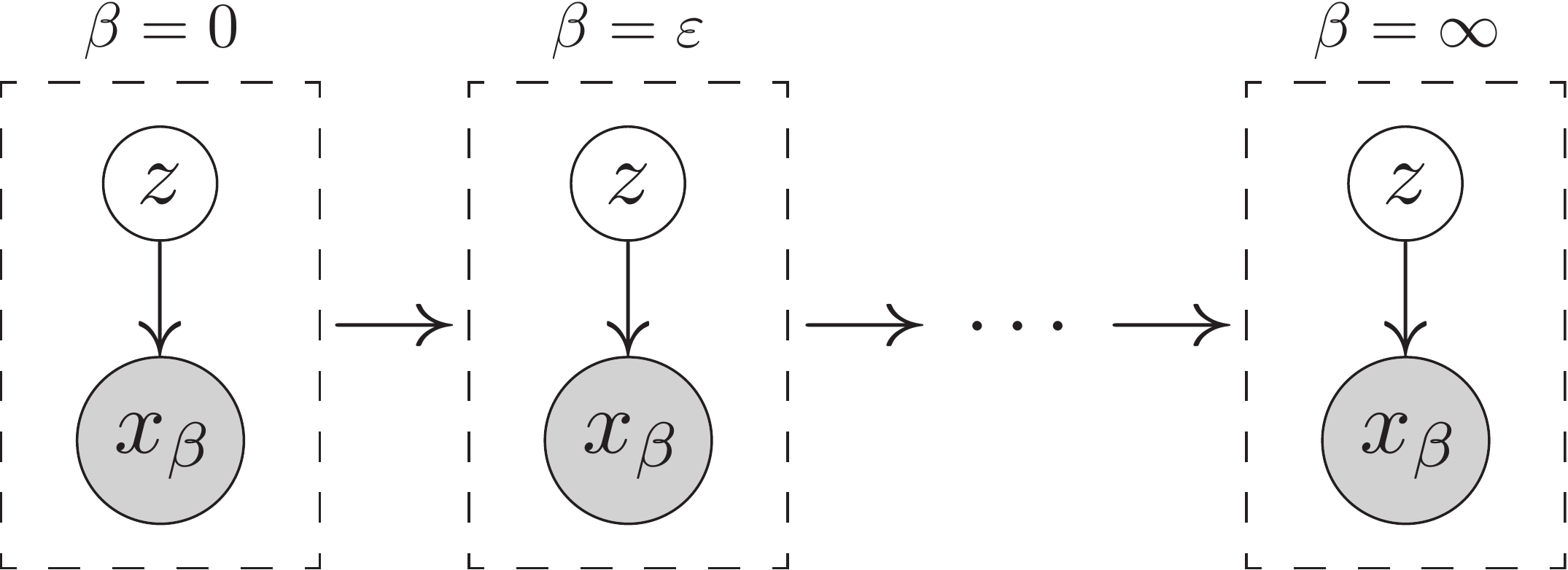}
\caption{{\it The schematic of $\beta$-GAN--- } GAN is initialized at $\beta=0$, corresponding to the uniform distribution. An annealing schedule is chosen to take $\beta$ from zero to infinity and the GAN training is performed simultaneously, where the parameters at each $\beta$ is initialized by the {\it optimal} parameters found at the previous smaller $\beta$. The notation $x_\beta$ refers to samples that come from $p_{{\rm data}}(x;\beta)$.}
\label{fig:annealedgan}
\end{figure}

With this setup, the minimax optimization task at each $\beta$ is:
\begin{eqnarray*}
\theta^*_G(\beta) &=& \argmin_{\theta_G} \max_{\theta_D} f(\theta_D,\theta_G;\beta),\\
f(\theta_D,\theta_G;\beta) &=& \expectation_{x\sim p_{{\rm data}}(x;\beta)} \log\left(D(x;\theta_D)\right) +  \expectation_{z\sim p(z)} \log(1-D(G(z;\theta_G);\theta_D)).
\end{eqnarray*}
Note that the optimal parameters $\theta^*_G$ and $\theta^*_D$ depend on $\beta$ implicitly. In $\beta$-GAN, the first task is to learn to sample the uniform distribution. It is then trained simultanously as the uniform distribution is smoothly annealed to the empirical distribution by increasing $\beta$. We chose a simple fixed geometric scheduling for annealing in this work. The algorithm is given below (see Fig.~\ref{fig:annealedgan} for the schematic):

\begin{algorithm*}[h!]
\caption{\small Minibatch stochastic gradient descent training of annealed generative adversarial networks. The inner loop can be replaced with other GAN architectures and/or other divergence measures. The one below uses the Jensen-Shannon formulation of Goodfellow {\it et al.} as the objective, as are all experiments in this paper.\comment{$\theta^*_{g, 0}$, $\theta^*_{d, 0}$ refers to the GAN equilibrium optimal parameters for generating the uniform distribution at $\beta=0$.}}
\begin{algorithmic}
\label{alg:AGAN}
\STATE{$\bullet$ Train GAN to generate uniform distribution and obtain $\theta^*_{g, 0}$ and $\theta^*_{d, 0}$.}
\STATE{$\bullet$ Receive $\beta_1$, $\beta_K$, and $K$, where $K$ is the number of cooling steps between/including $\beta_1$ and $\beta_K$.}
\STATE{$\bullet$ Compute $\alpha>1$ as the geometric cooling factor:
	 \[
\alpha= \left({\frac{\beta_K}{\beta_1}}\right)^\frac{1}{K}
\]
}

\STATE{$\bullet$ Initialize $\beta$: $\beta \leftarrow \beta_1$}
\STATE{$\bullet$ Initilize $\theta_{g, \beta} \leftarrow \theta^*_{g, 0}$ and $\theta_{d, \beta} \leftarrow \theta^*_{d, 0}$}
\FOR{number of cooling steps $(K)$}
\FOR{number of training steps $(n)$}
\STATE{$\bullet$ Sample minibatch of $m$ noise samples $\{ z^{(1)}, \dots, z^{(m)} \}$ from noise prior $p(z)$.}
\STATE{$\bullet$ Sample minibatch of $m$ examples $\{ x^{(1)}, \dots, x^{(m)} \}$ from data generating distribution $p_{{\rm data}}(x;\beta)$.}
\STATE{$\bullet$ Update the discriminator by ascending its stochastic gradient:
\[
\nabla_{\theta_{d, \beta}} \frac{1}{m} \sum_{i=1}^m \left[
\log D\left(x^{(i)};\theta_{d, \beta}\right)
+ \log \left(1-D\left(G\left(z^{(i)};\theta_{g, \beta}\right);\theta_{d, \beta}\right)\right)
\right].
\]}

\STATE{$\bullet$ Sample minibatch of $m$ noise samples $\{ z^{(1)}, \dots, z^{(m)} \}$ from noise prior $p(z)$.}
\STATE{$\bullet$ Update the generator by descending its stochastic gradient:
\[
\nabla_{\theta_{g, \beta}} \frac{1}{m} \sum_{i=1}^m
\log \left(1-D\left(G\left(z^{(i)};\theta_{g, \beta}\right);\theta_{d, \beta}\right)\right)
.
\]}
\ENDFOR
\STATE{$\bullet$ Increase $\beta$ geometrically: $\beta \leftarrow \beta*\alpha$
}
\ENDFOR
\STATE{$\bullet$ Switch from $p_{data}(x;\beta_K)$ to the empirical distribution ($\beta=\infty$) for the final epochs.}
\end{algorithmic}
\end{algorithm*}

The convergence of the algorithm is based on the following conjecture:

{\it In the continuous annealing limit from the uniform distribution to the data distribution GAN remains stable at the equilibrium, assuming G and D have large capacity and that they are initialized at the minimax equilibrium for generating the uniform distribution\footnote{This requires $\dim(z) \geq d$.} in the ambient space $\mathcal{X}$.}

\clearpage

\begin{figure}[h!]

\begin{tabular}{  p{4cm}  p{4cm}  p{4cm} }
\vspace{-0.7cm}
\includegraphics[width=5cm, height=5cm]{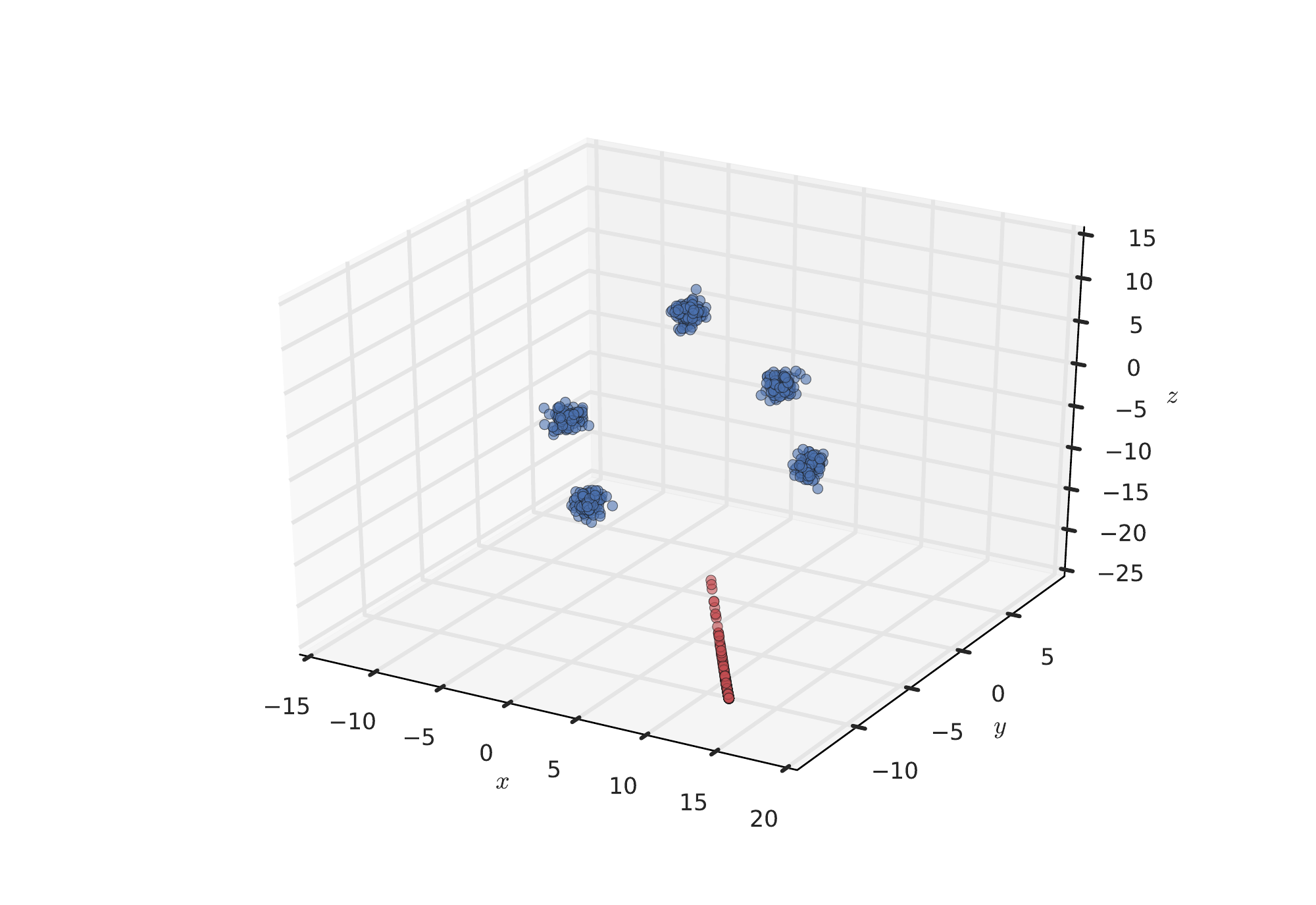}
&
\vspace{-0.7cm}
\includegraphics[width=5cm, height=5cm]{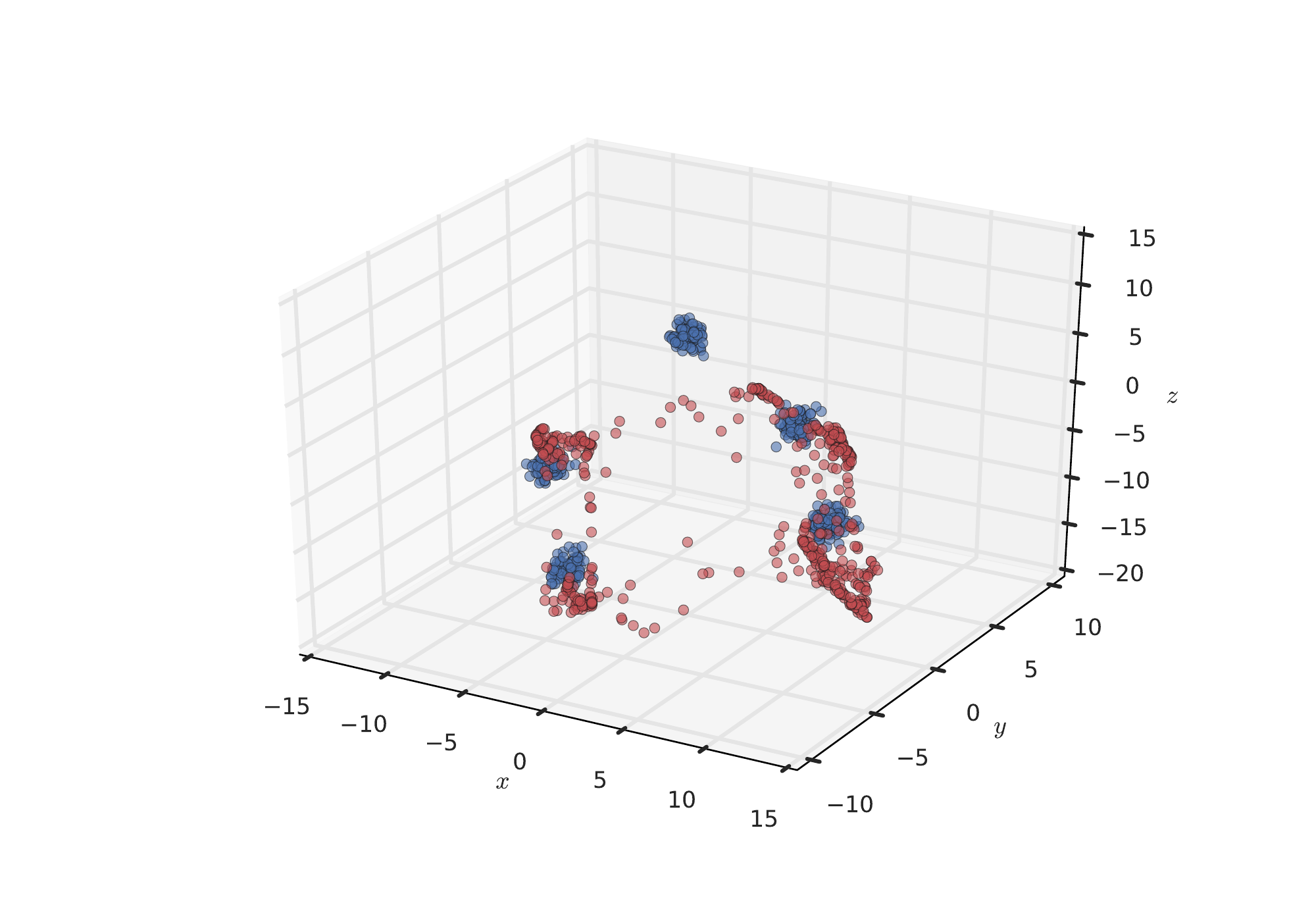}
&
\vspace{-0.7cm}
\includegraphics[width=5cm, height=5cm]{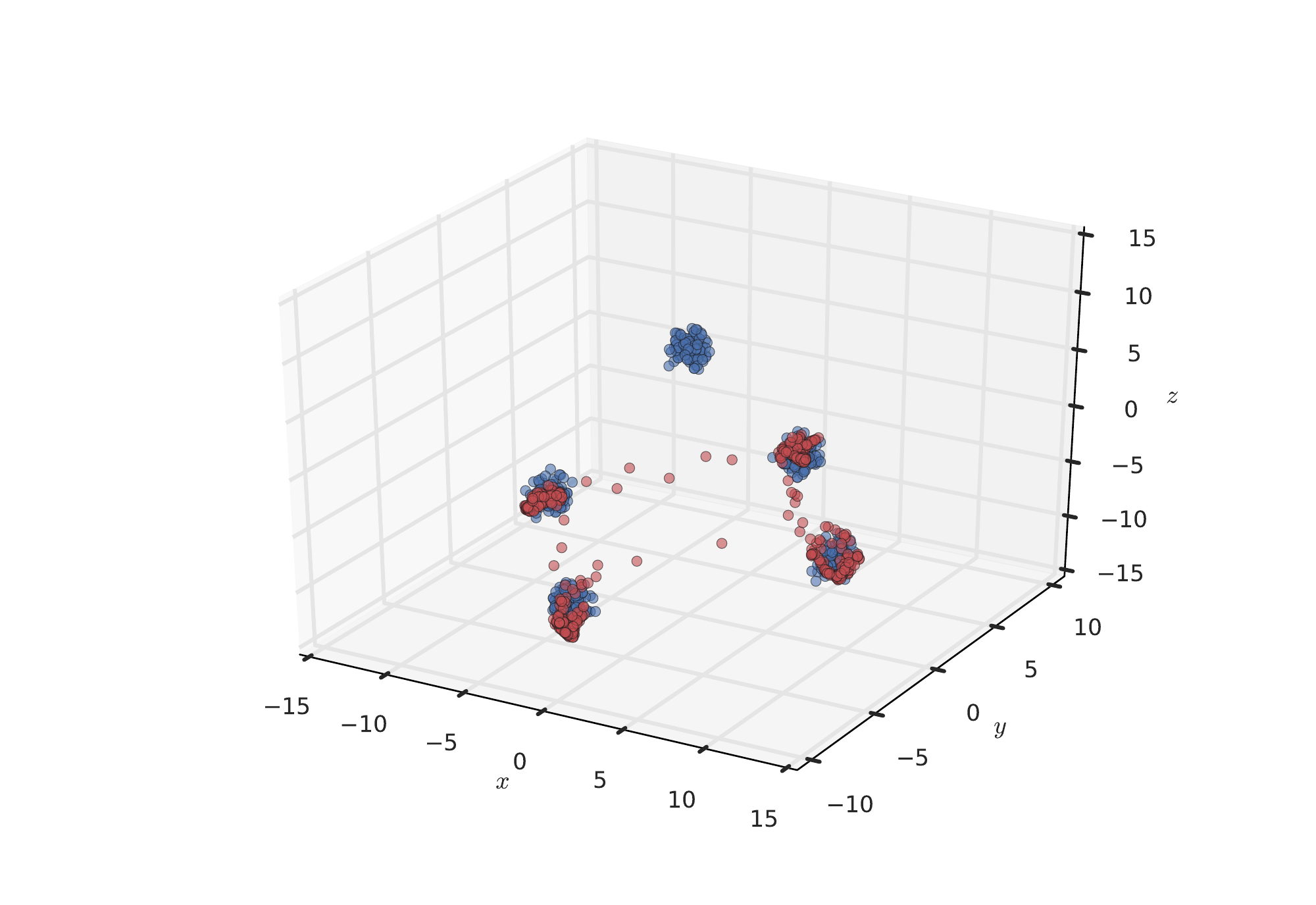}
\\

{\vspace{-0.5cm} \hspace{1.5cm} \centering \small $ \tau={\tt 1e3}$ \par}
&
\vspace{-0.5cm}
\hspace{1.4cm}
{\centering \small $\tau={\tt 2e3}$ \par}
&
\vspace{-0.5cm}
\hspace{1.4cm}
{\centering \small $\tau={\tt1e5}$ \par}
\vspace{-2cm}
\\
\vspace{-1cm}
\includegraphics[width=5cm, height=5cm]{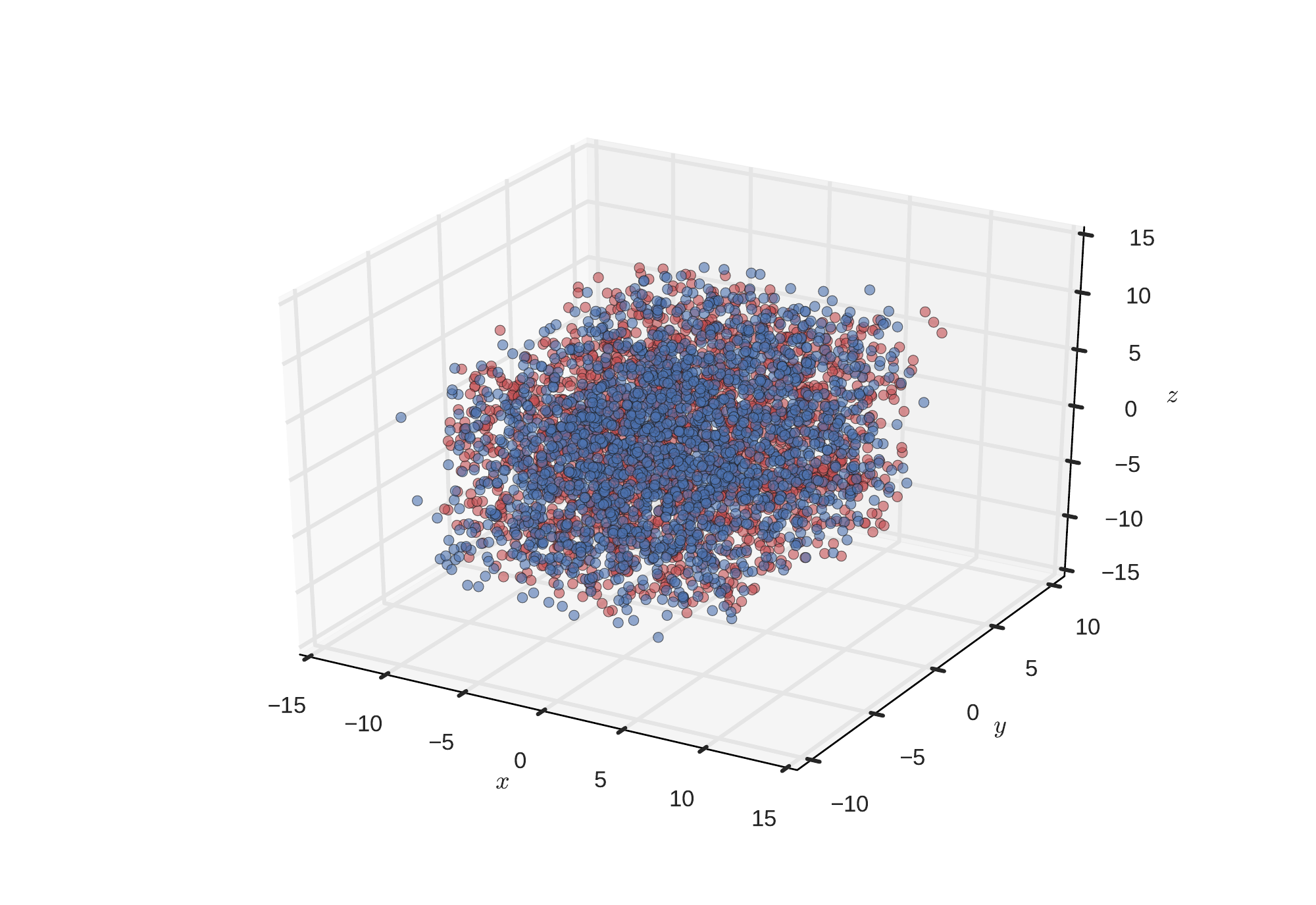}
&
\vspace{-1cm}
\includegraphics[width=5cm, height=5cm]{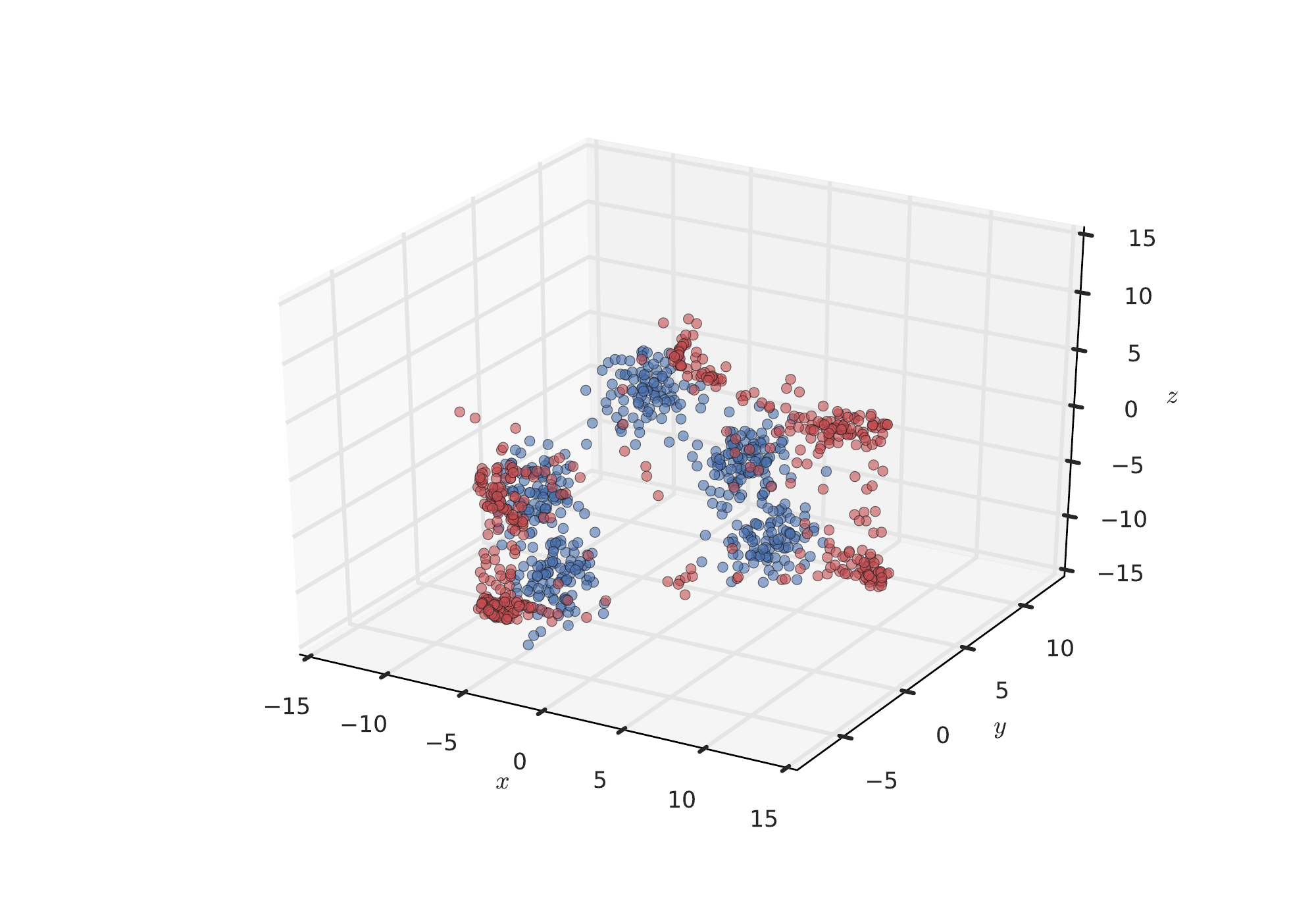}
&
\vspace{-1cm}
\includegraphics[width=5cm, height=5cm]{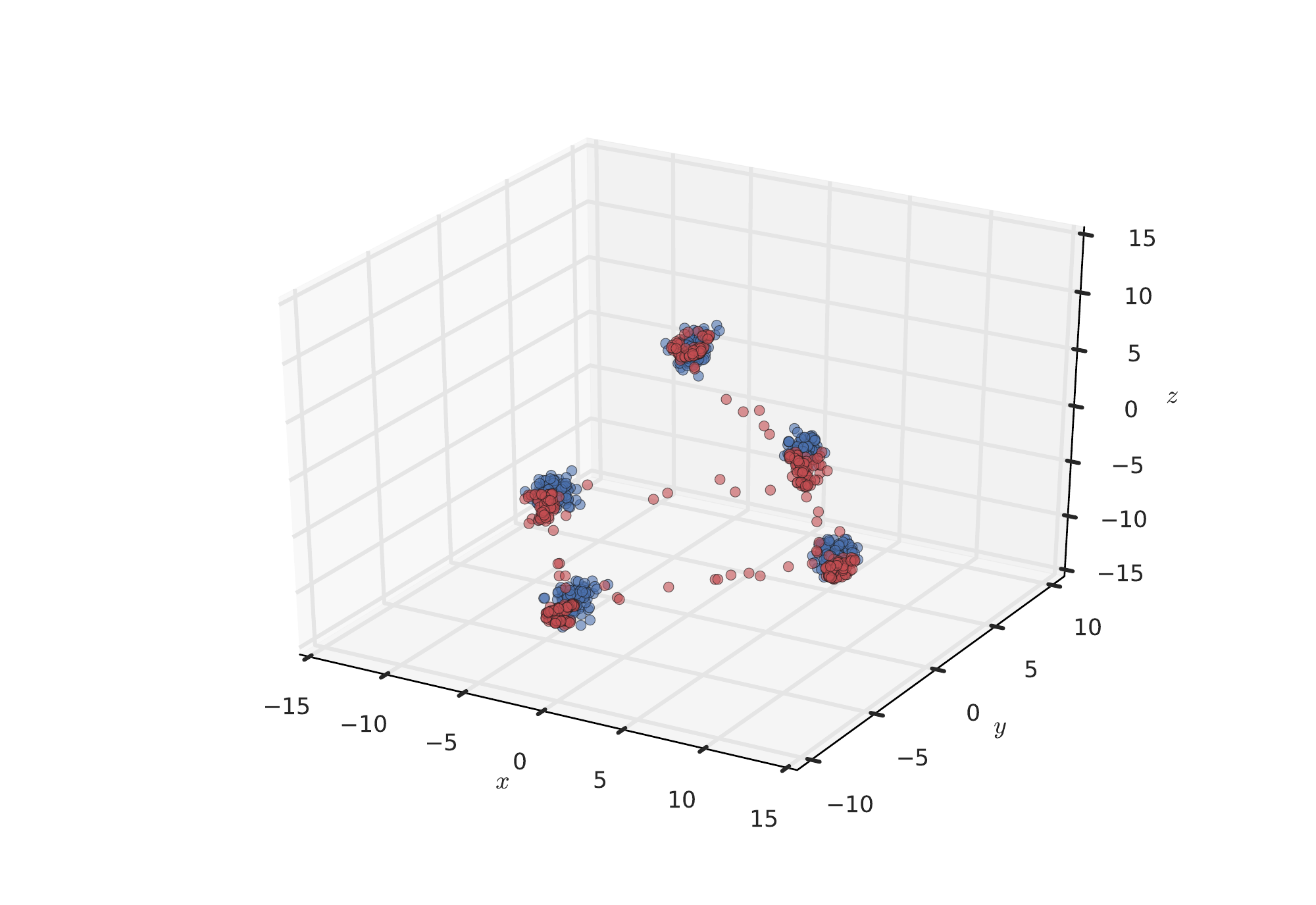}
\\
\vspace{-0.6cm}
\hspace{1.4cm}
{\centering \small $\beta={\tt 0}, \tau={\tt1e3}$ \par}
&
\vspace{-0.6cm}
\hspace{1.4cm}
{\centering \small $\beta={\tt 1}, \tau={\tt2e3}$ \par}
	 &
\vspace{-0.6cm}
	   \hspace{1.4cm}
	{\centering \small $\beta={\tt 10}, \tau={\tt1e4}$ \par}
\\
\vspace{-1cm}
\includegraphics[width=5cm, height=5cm]{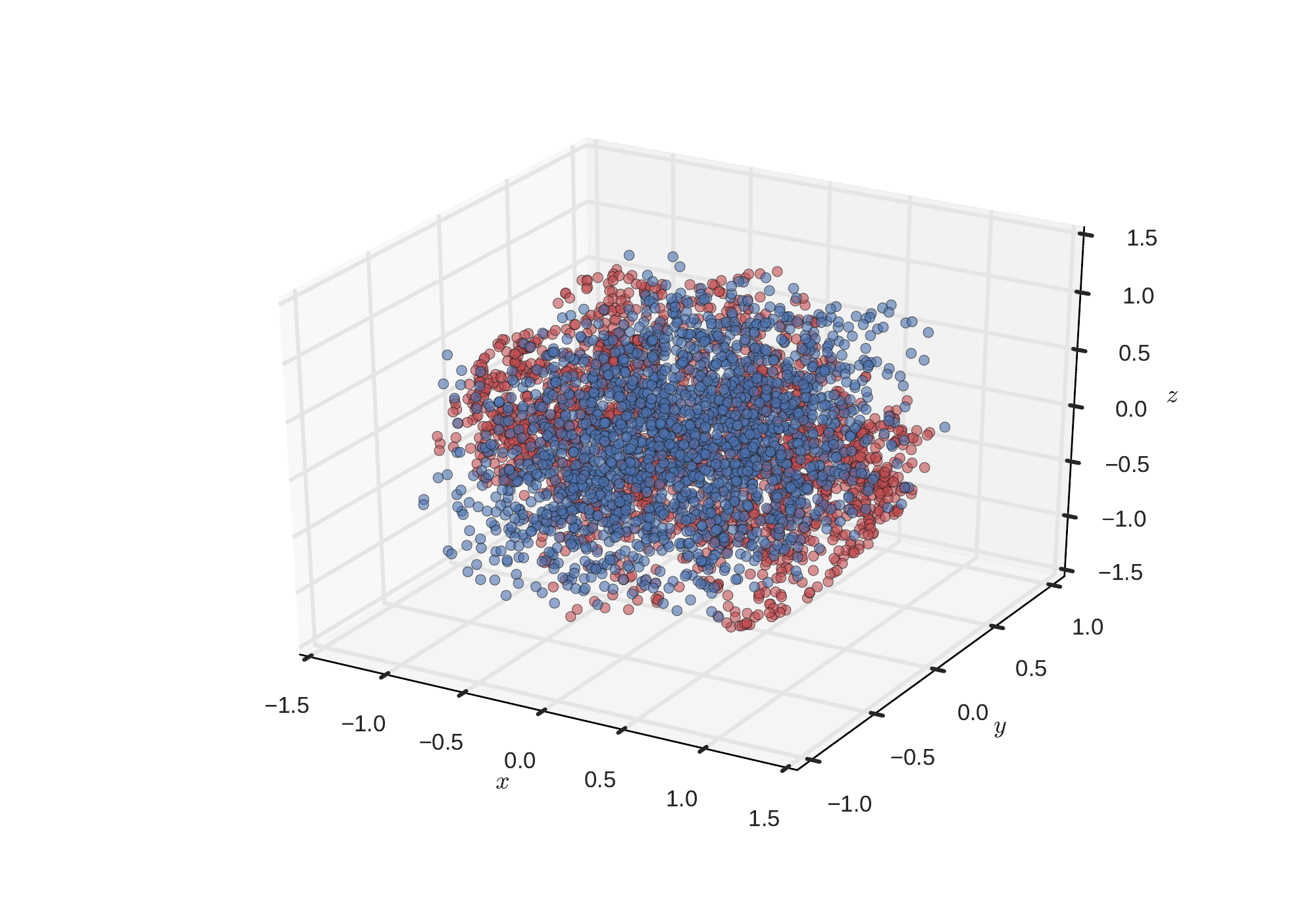}
&
\vspace{-1cm}
\includegraphics[width=5cm, height=5cm]{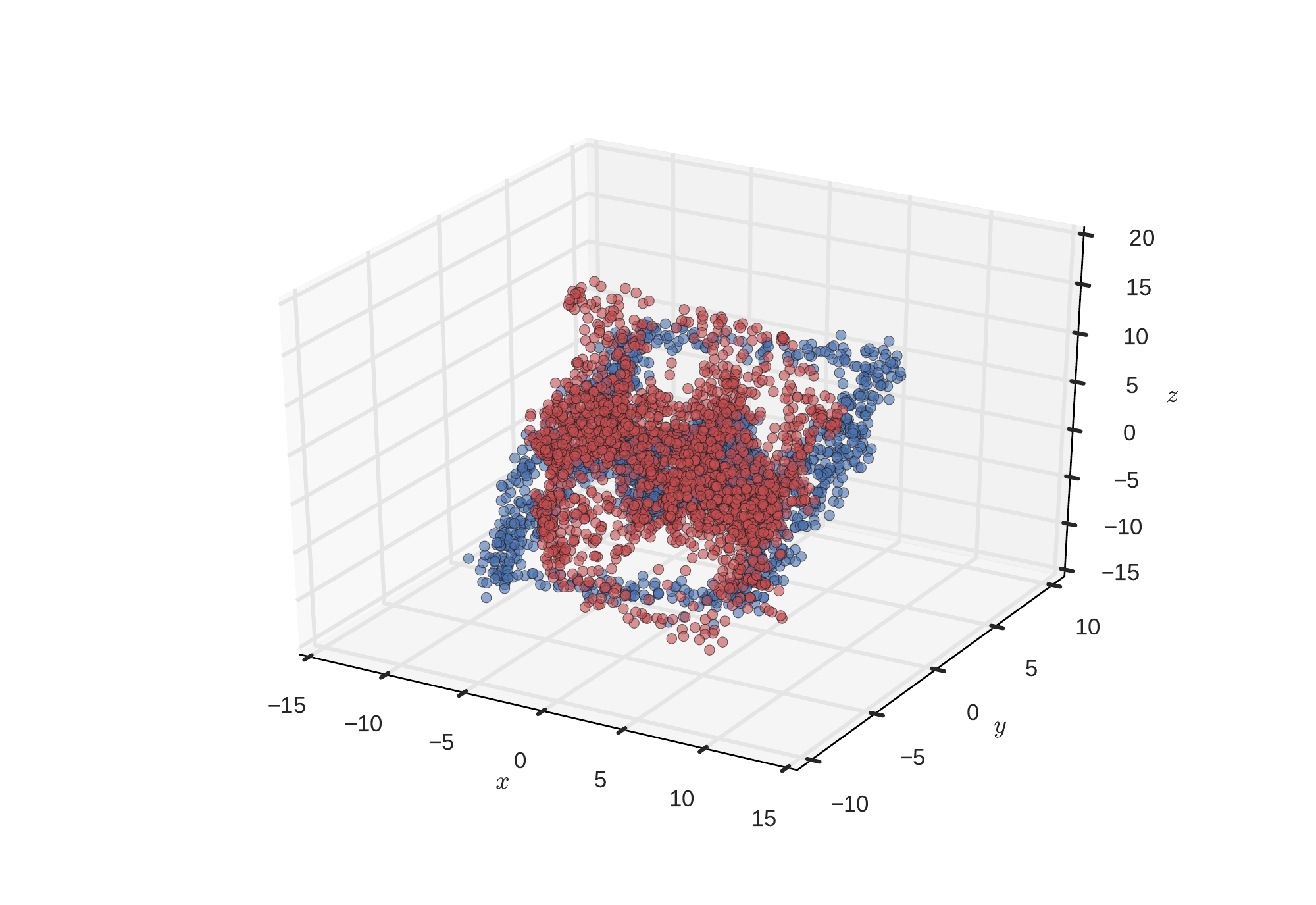}
&
\vspace{-1cm}
\includegraphics[width=5cm, height=5cm]{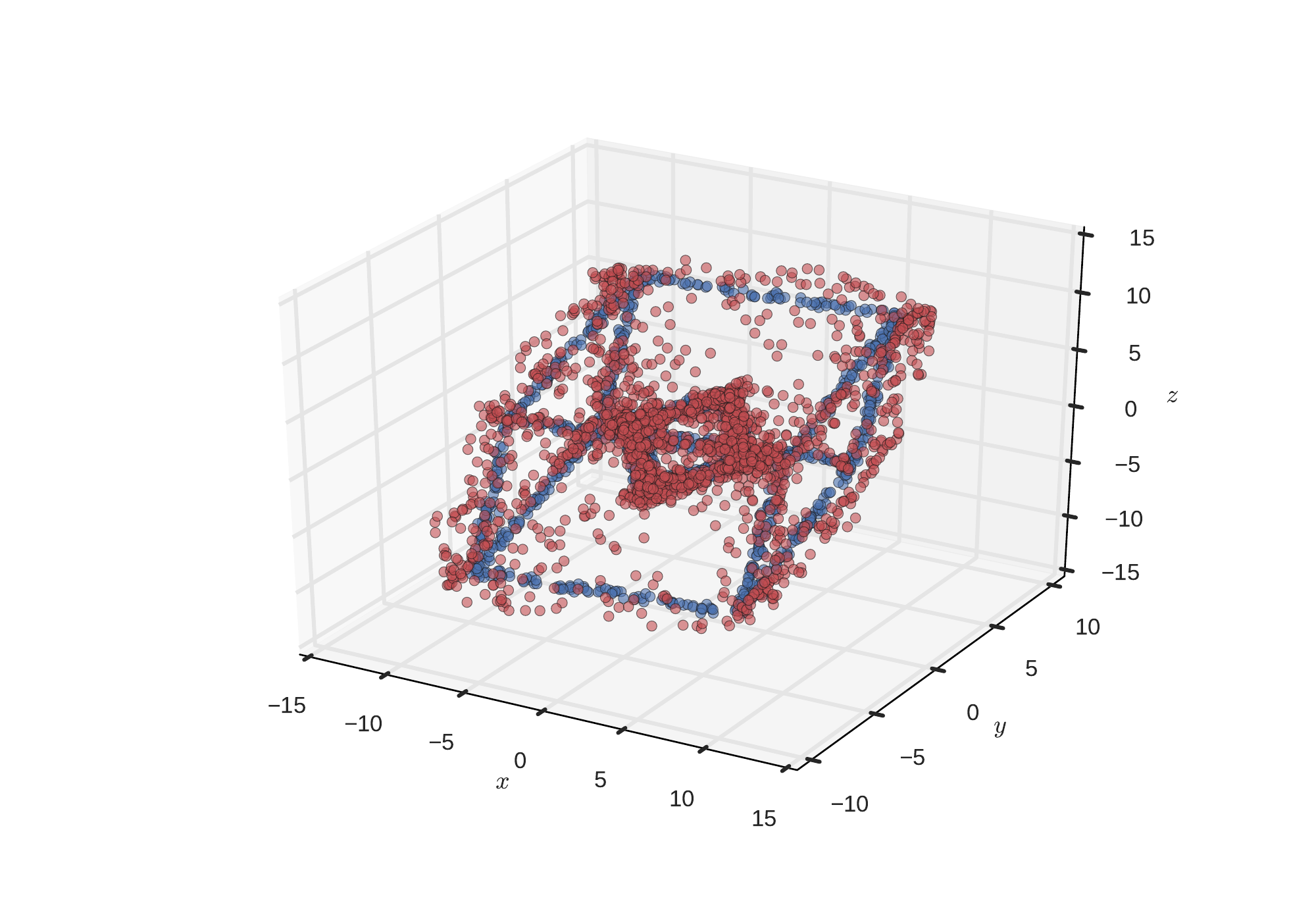}

\\
\vspace{-1cm}
\hspace{1.4cm}
{\centering \small $\beta={\tt 0}, \tau={\tt1e3}$ \par}
&
\vspace{-1cm}
\hspace{1.4cm}
{\centering \small $\beta={\tt 1}, \tau={\tt2e3}$ \par}
	 &
	 \vspace{-1cm}
	   \hspace{1.4cm}
	{\centering \small $\beta={\tt 10}, \tau={\tt1e4}$ \par}

\end{tabular}
\vspace{-0.5cm}
\caption{{\it Three dimensional example --- }
{\it Top row:} The performance of vanilla GAN on a mixture of five Gaussian components in three dimensions. {\it Middle row:} The performance of $\beta$-GAN on the same dataset. {\it Bottom row:} The performance of $\beta$-GAN on the synthesized mixture of two cubes. Blue/red dots are real/generated data. To compare the computational cost, we report $\tau$, which is the total number of gradient evaluations from the start. We use the architecture G:[z(3) | ReLU(128) | ReLU(128) | Linear(3)] and D:[x(3) | Tanh(128) | Tanh(128) | Tanh(128) | Sigmoid(1)] for generator and discriminator where the numbers in the parentheses show the number of units in each layer. The annealing parameters are [$\beta_1=0.1,\ \beta_K=10,\ K=20$].}
\label{fig:Results}
\end{figure}

\section{Experiments} \label{sec:EXPS}
\begin{figure}[h!]
	\centering
	\subfigure[Mode collapse to frozen noise]{
		\includegraphics[width=0.35\linewidth]{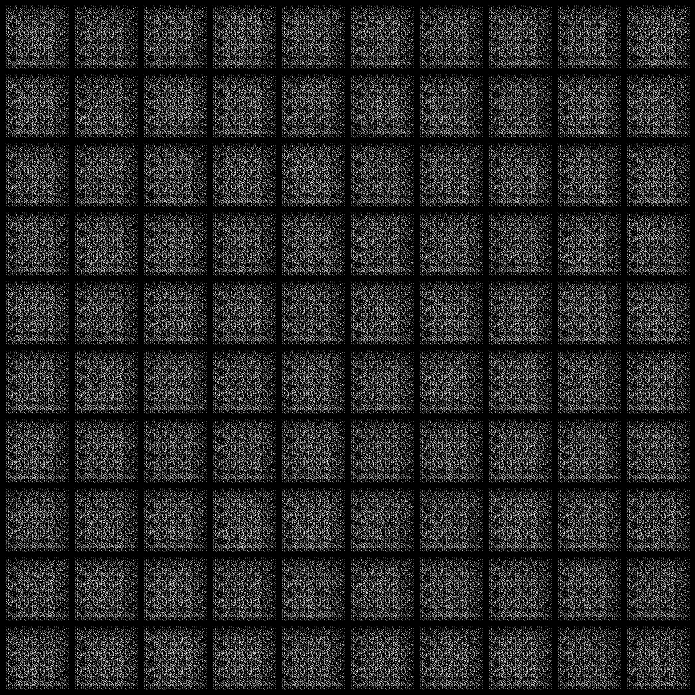}\label{fig:frozen_noise}}
	\subfigure[Samples from uniform distribution]{
		\includegraphics[width=0.35\linewidth]{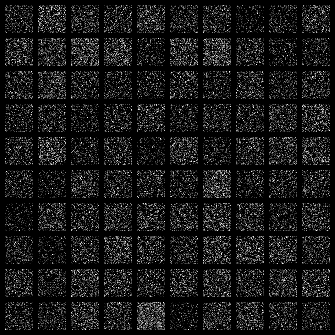}\label{uniform_noise}}
\caption{{\it Uniform distribution generation performance ---} (a) The frozen noise pattern that we observe in our training using smooth nonlinearities in the generative network. The result here is for Tanh. (b) The mode collapse to frozen noise was resolved using piece-wise linear ReLU activation in the generator.}
	\label{fig:frozen_uniform_noise}
	\vspace{-0.2cm}
\end{figure}

$\beta$-GAN starts with learning to generate the uniform distribution in the ambient space of data. The mapping that transforms the uniform distribution\footnote{We used the uniform prior for $z$ in all our experiments.} to the uniform distribution of the same dimension is an affine function. We therefore used only ReLU nonlinearity in the generative network to make the job for the generator easier. The performance of the network in generating the uniform distribution was degraded by using smooth nonlinearities like Tanh. It led to immediate mode collapse to \emph{frozen noise} instead of generating high-entropy noise (see Figure~\ref{fig:frozen_uniform_noise}). The mode collapse to frozen noise was especially prominent in high dimensions.

\subsection{Toy examples} \label{subsec:TEXM}

\begin{figure}[h!]
	\centering
	\subfigure[$\beta$-GAN for MoG with 5 modes]{
		\includegraphics[width=0.49\linewidth]{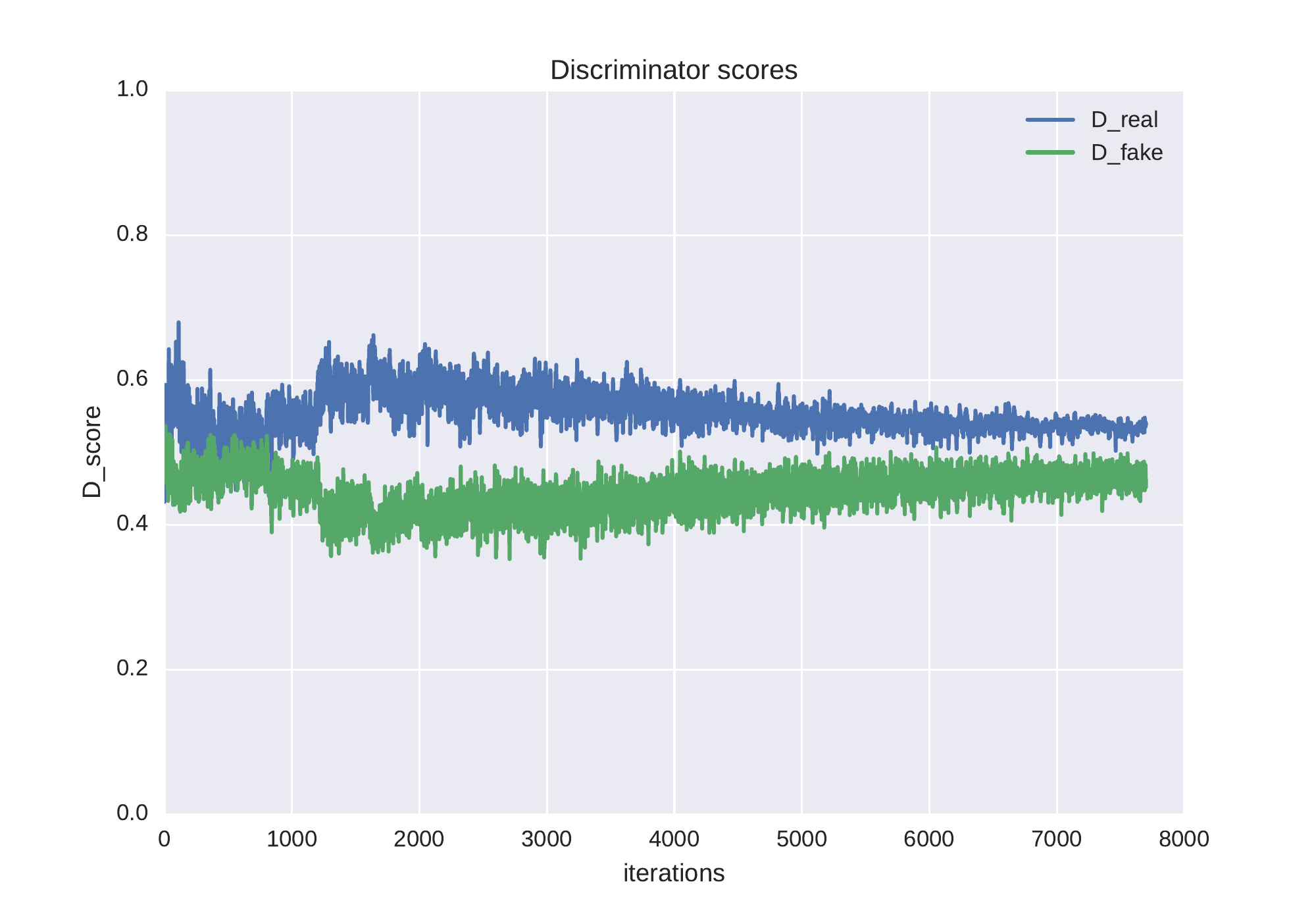}}
	\subfigure[$\beta$-GAN for MoG with 10 modes]{
		\includegraphics[width=0.49\linewidth]{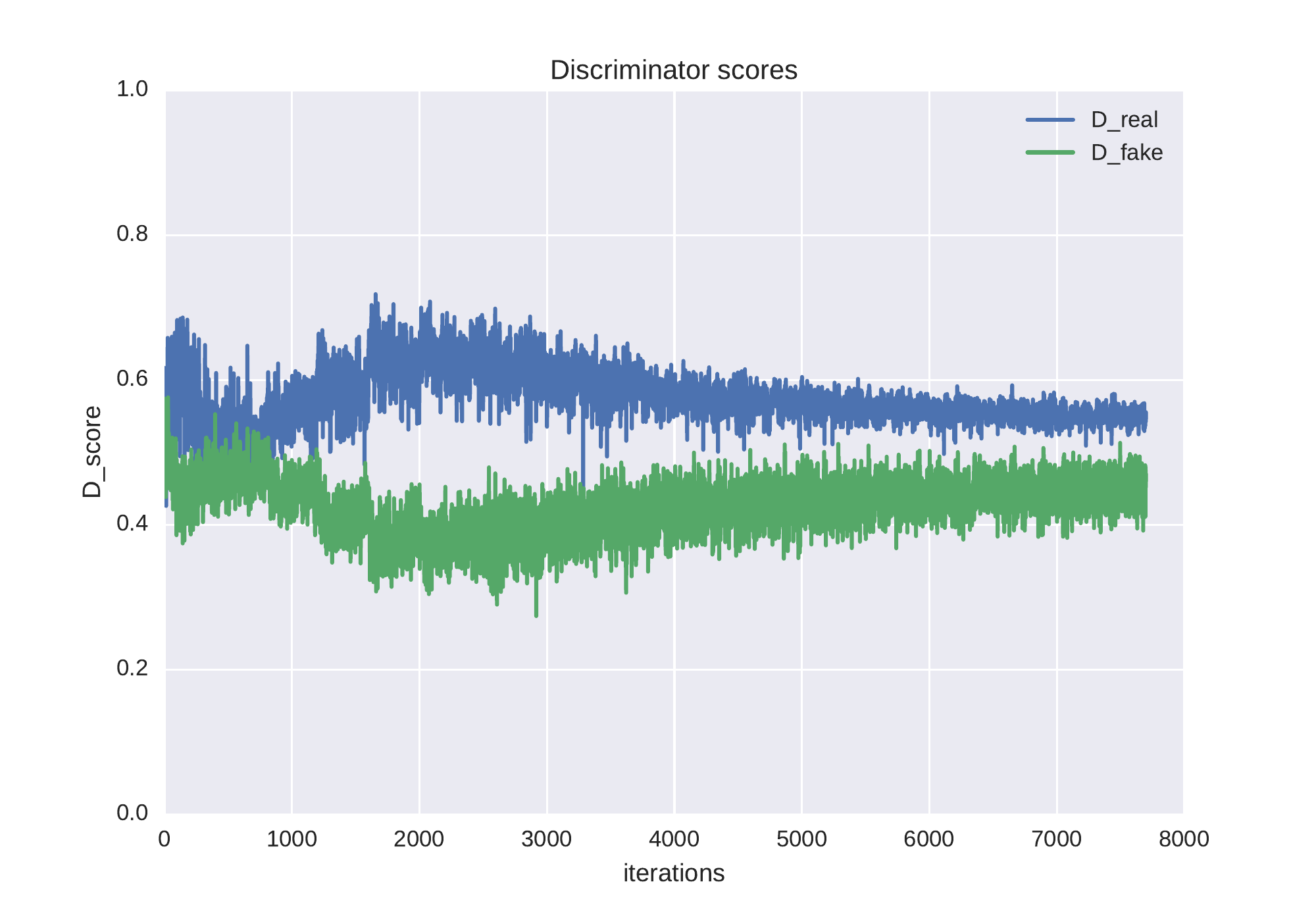}}
	\subfigure[Vanilla GAN for MoG with 5 modes]{
		\includegraphics[width=0.49\linewidth]{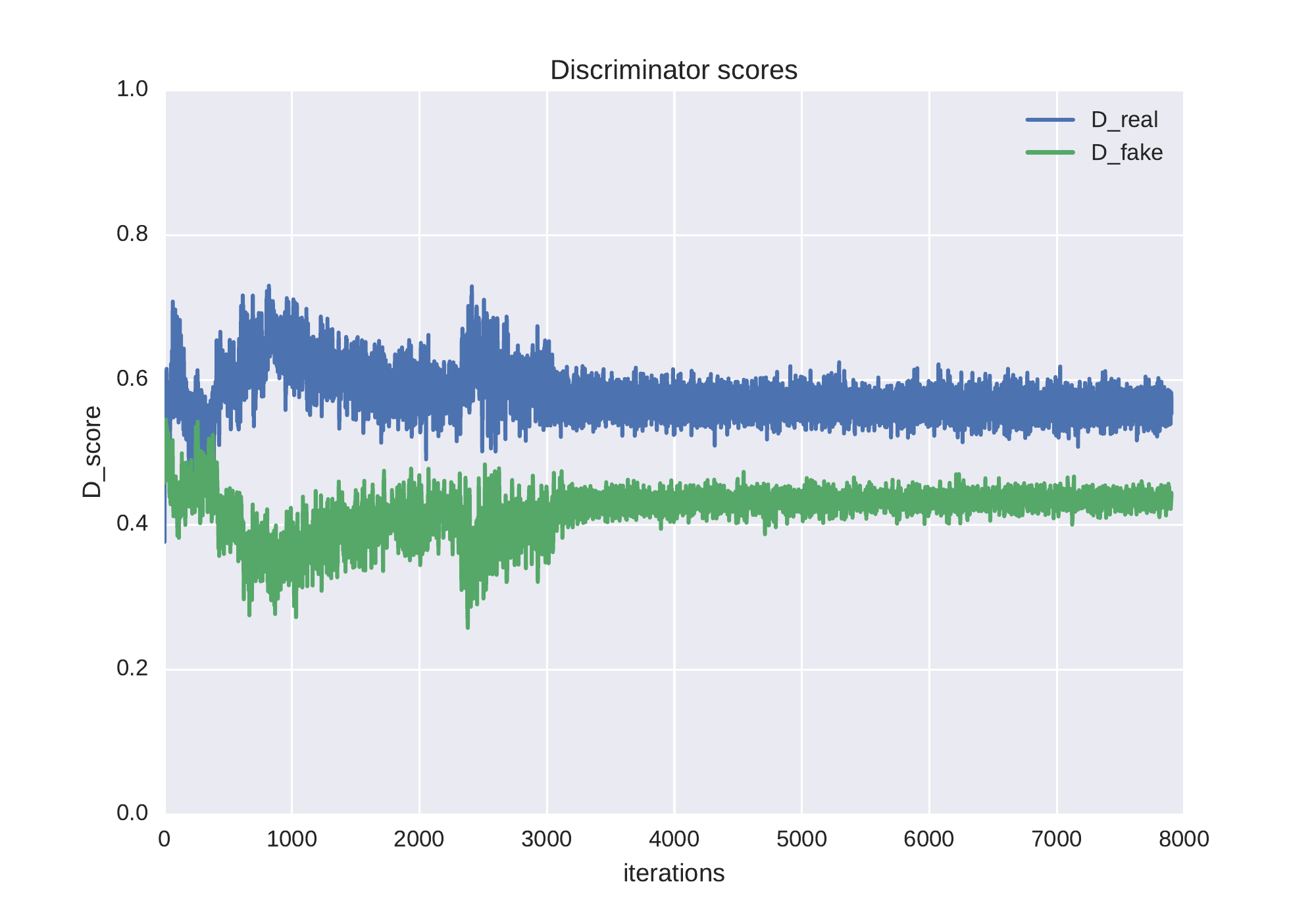}}
 \subfigure[Vanilla GAN for MoG with 10 modes]{
		\includegraphics[width=0.49\linewidth]{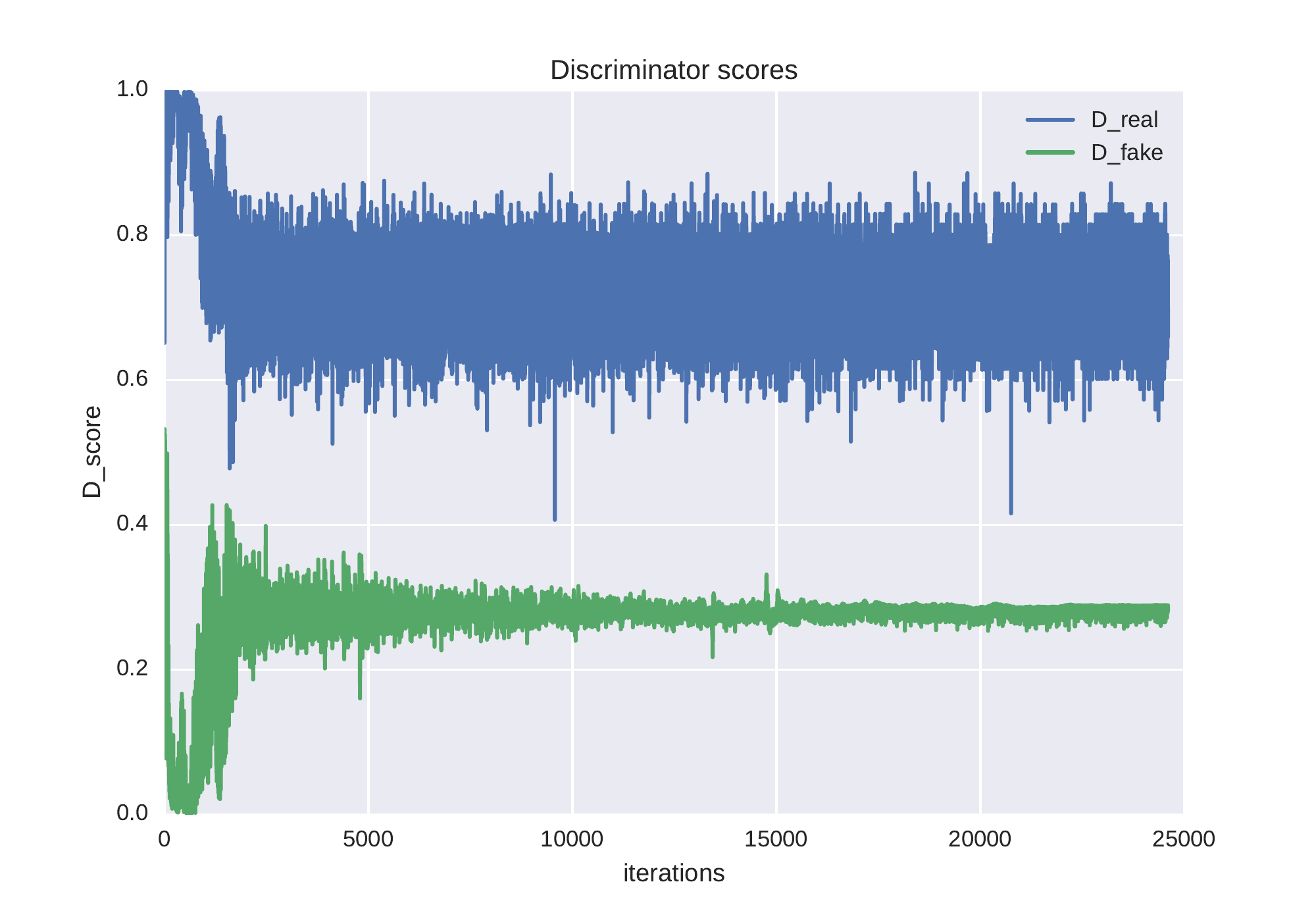}}
	\caption{ {\it Training curves ---}
The curves shown here are the output of the discriminator (which is a classifier in this case) for the real and generated samples. For $\beta$-GAN the training curves show a more stable behavior with more robustness to the complexity of input data (a,b). However, when the data gets more complex,  vanilla GAN performance gets worse signified the growing gap between $D_{\rm real}$ and $D_{\rm fake}$ (c,d).}
	\label{fig:training_curves}
\end{figure}

To check the stability of $\beta$-GAN, we ran experiments on mixtures of 1D, 2D, 3D Gaussians,  and a mixture of two cubic frames in 3D. The 3D results are presented here. The reported results for vanilla GAN (top row of Fig.~\ref{fig:Results}) was the best among many runs; in most experiments vanila-GAN captured only one mode or failed to capture any mode. However, $\beta$-GAN produced similar results consistently. In addition, vanilla GAN requires the modification of the generator loss to $\log(D(G(z;\theta_G)))$ to avoid saturation of discriminator~\citep{goodfellow2014generative}, while in $\beta$-GAN we did not make any modification, staying with the generator loss $\log(1-D(G(z;\theta_G);\theta_D))$. In the experiments, the total number of training iterations in $\beta$-GAN was the same as vanilla GAN, but distributed over many intermediate temperatures, thus curbing the computational cost. We characterized the computation cost by the total number of gradient evaluations $\tau$ reported in the Fig.~\ref{fig:Results}. We also compared the training curves of $\beta$-GAN and vanilla GAN for mixtures of five and ten Gaussians  (see Fig.~\ref{fig:training_curves}).

We also synthesized a dataset that is a mixture of two cubic frames, one enclosed by the other. This dataset is interesting since the data is located on disjoint 1D manifolds within the 3D ambient space. $\beta$-GAN performs well in this case in every run of the algorithm (see bottom row of Fig.~\ref{fig:Results})

We should emphasize that different GAN architectures can be easily augmented with $\beta$-GAN as the outer loop. In the 3D experiments here, we chose the original architecture of generative adversarial network from ~\citep{goodfellow2014generative} as the inner loop (see Algorithm~\ref{alg:AGAN}). In the next section we show the results for more sophisticated GAN architectures.

\subsection{High-dimensional examples}\label{subsec:HDEXM}

To check the performance of our method in higher dimensions we applied $\beta$-GAN to the  MNIST dataset~\citep{lecun1998gradient} with the dimension $28\times 28$ and CelebA dataset~\citep{liu2015faceattributes} with the the dimension $64\times 64 \times 3$. Once again, we start from generating the uniform distribution in the ambient space of the data and we use only piecewise linear activation functions for the generative network due to the \emph{frozen noise} mode collapse that we discussed earlier.

\begin{figure}[h!]
	\centering
	\subfigure[Generated samples for $\beta=0.1$]{
		\includegraphics[width=0.35\linewidth]{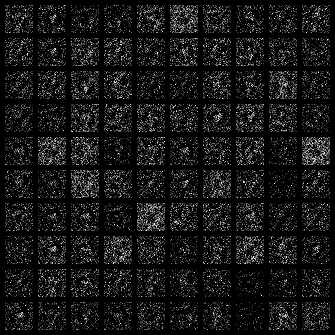}}
	\subfigure[Generated samples for $\beta=1$]{
		\includegraphics[width=0.35\linewidth]{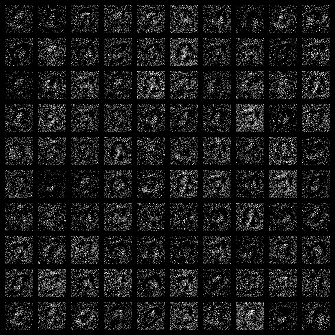}}
	\subfigure[Generated samples for $\beta=5$]{
		\includegraphics[width=0.35\linewidth]{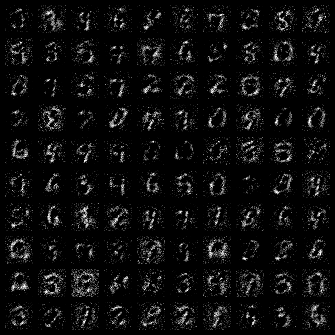}}
 \subfigure[Generated samples for $\beta=\infty$]{
		\includegraphics[width=0.35\linewidth]{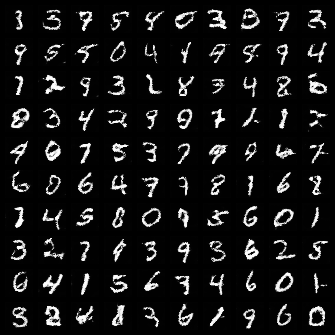}}
	\caption{{\it $\beta$-GAN trained on MNIST with $\dim(z)=28\times28$ ---} Samples generated from MNIST during annealing procedure. The network starts from generating the uniform distribution ${\rm Uniform}[-1,1]^{28\times28}$ and gradually generates samples corresponding to each value of $\beta$. We use the fullly connected architecture G:[z(784) | BNReLU(256) | BNReLU(256) | BNReLU(256) | Linear(784)] and D:[x(784) | BNReLU(256) | BNReLU(512) | BNReLU(512) | Sigmoid(1)] for generator and discriminator where the numbers in the parentheses show the number of units in each layer. BNReLU is batch normalization~\citep{IoffeS15batchnormal} concatenated with ReLU activation. The annealing parameters are [$\beta_1=0.1,\ \beta_K=10,\ K=20$] the same as 3D experiment in Fig.~\ref{fig:Results}.}
	\label{fig:annealing_mnist}
	\vspace{-0.2cm}
\end{figure}

\begin{figure}[t!]
	\centering
	\subfigure[Generated samples for $\beta=0.1$]{
		\includegraphics[width=0.49\linewidth]{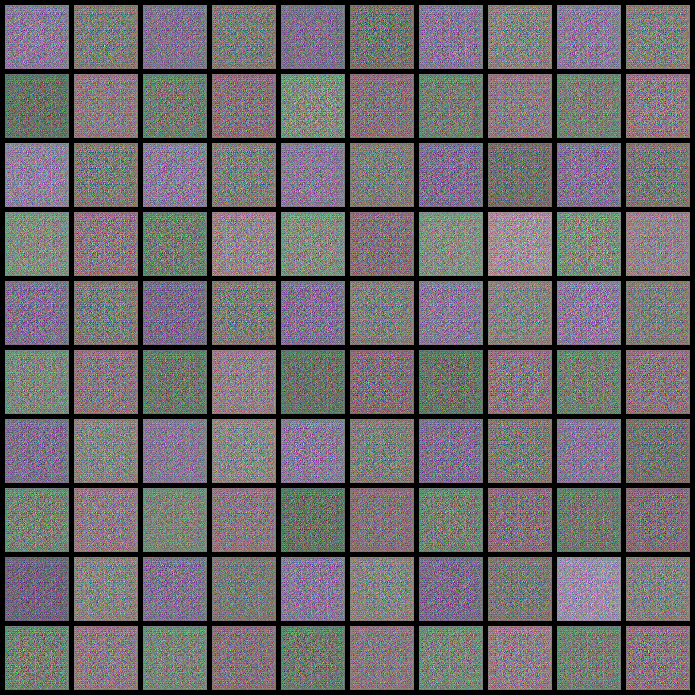}}
	\subfigure[Generated samples for $\beta=1$]{
		\includegraphics[width=0.49\linewidth]{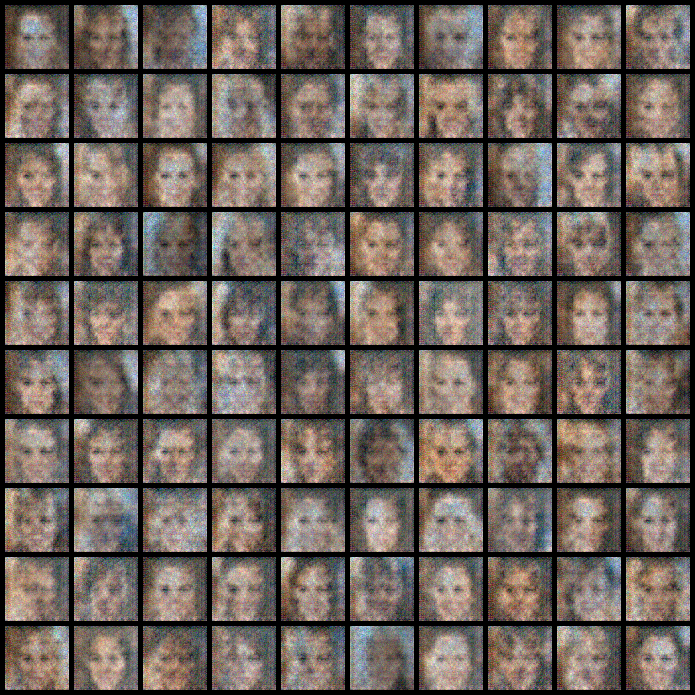}}
	\subfigure[Generated samples for $\beta=5$]{
		\includegraphics[width=0.49\linewidth]{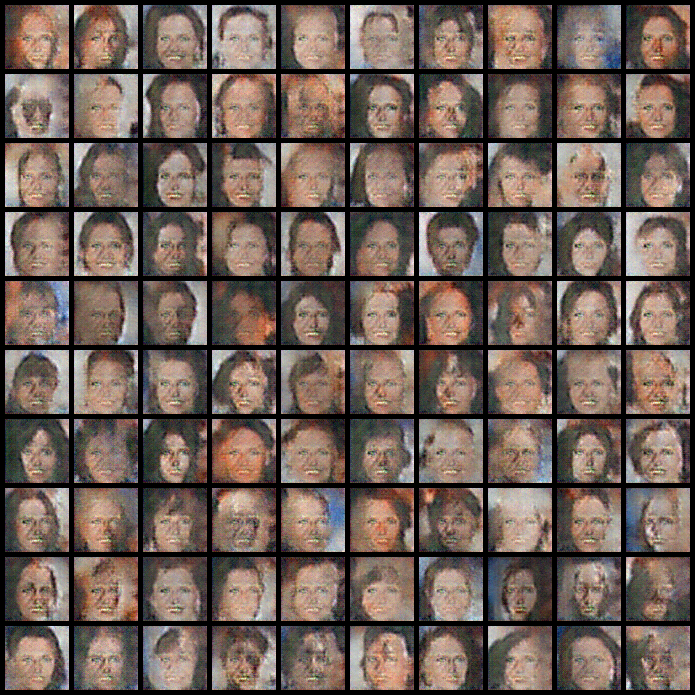}}
 \subfigure[Generated samples for $\beta=\infty$]{
		\includegraphics[width=0.49\linewidth]{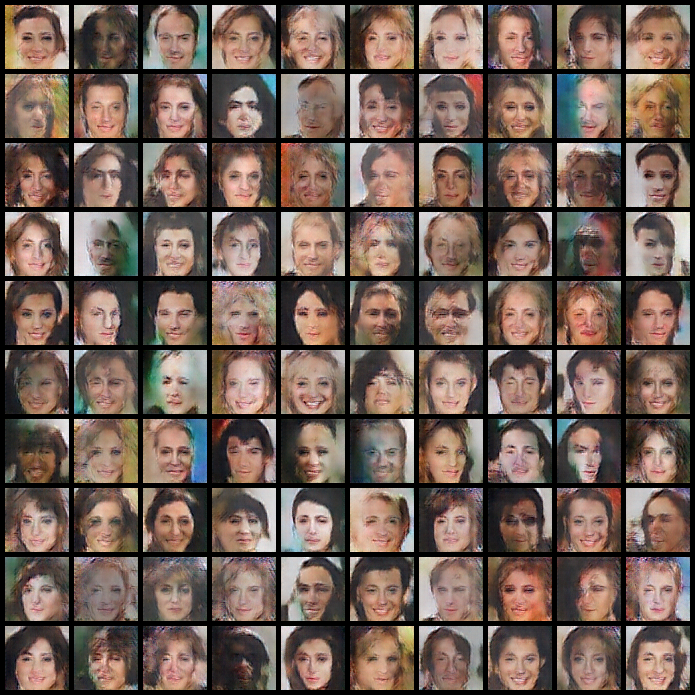}}
	\caption{{\it $\beta$-GAN trained on CelebA with $\dim(z)=64 \times64 \times 3$ ---} Samples generated from CelebA dataset during annealing procedure. The network starts from generating the uniform distribution ${\rm Uniform}[-1,1]^{64\times64\times3}$ and gradually generates samples corresponding to each value of $\beta$. We borrowed DCGAN architecture from ~\citep{radford2015unsupervised} except that the input noise of the generative network has the dimension of data and the output layer is changed to linear instead of Tanh. The annealing parameters are [$\beta_1=0.1,\ \beta_K=10,\ K=20$] the same as 3D experiment in Fig.~\ref{fig:Results}.}
	\label{fig:annealing_celeb}
	\vspace{-0.2cm}
\end{figure}

The performance of $\beta$-GAN for the MNIST dataset with a fully connected network is shown in Fig..~\ref{fig:annealing_mnist}. As $\beta$ gradually increases, the network learns to generate noisy images corresponding to each temperature. The results converge to clean MNIST images in the last epochs of training, where data distribution is cooled down at high value of $\beta$. Also during intermediate epochs, noisy digits are generated, which are still diverse. This behavior is in contrast with the training of vanilla GAN, where collapsing at single mode is common in intermediate iterations. The same experiment was performed for CelebA dataset with the same annealing procedure, starting from the uniform distribution and annealing to the data distribution. The results are reported in Figure~\ref{fig:annealing_celeb}.

Regarding annealing from the uniform distribution to the data distribution, we used the same annealing schedule in all our experiments -- for mixture of Gaussians (different number of modes), mixture of interlaced cubes, MNIST and CelebA -- and we consistently achieved the results reported here. This highlights the stability of $\beta$-GAN. We think this stability is due to the $\beta$-GAN conjecture (see Section~\ref{sec:MODEL}) even though the annealing is not continuous in the experiments. 

We emphasize that both MNIST and CelebA images were generated with $\dim(z) = 28\times 28$ and $\dim(z) = 64\times 64\times 3$, the dimensions of their ambient space respectively. At the beginning, the support of the generated distribution (i.e. the uniform distribution) is the ambient space. $\beta$-GAN learns during annealing, step-by-step, to shape the space-filling samples to samples that lie on the manifold of MNIST digits and CelebA faces.  

\section{Discussion}
In this work, we took a departure from the current practices in training adversarial networks by giving the generative network the capacity to fill the ambient space in the form of the uniform distribution. The uniform distribution was motivated from statistical mechanics, where we imagined the data particles diffusing like ink dropped in water. The parameter $\beta$ can be thought of as a surrogate for this diffusion process. There are in fact many ways to transform the data distribution to the uniform distribution. An approach that is non-Gaussian is  flipping bits randomly in the bit representation~\citep{saremi2013hierarchical, saremi2016correlated} -- this process will take any distribution to the uniform distribution in the limit of many bit flips. The starting point in $\beta$-GAN has deep consequences for the adversarial training. It is a straightforward solution to the theoretical problems raised in~\citep{arjovsky2017towards}, since the results there were based on $\dim(z) < d$. However, despite $\beta$-GAN's success in our experiments, the brute force $\dim(z)\geq d$ may not be practical in large dimensions. We are working on ideas to incorporate multi-scale representations~\citep{denton2015deep} into this framework, and are considering dimensionality reduction as a \say{pre-processing} step before feeding data into $\beta$-GAN. To emphasize the robustness of $\beta$-GAN, we reported results with a fixed annealing schedule, but we have also explored ideas from feedback control~\citep{BerthelotSM17BEGAN} to make the annealing adaptive.     



\clearpage

\subsection*{Acknowledgments}
SS acknowledges the support by CIFAR. We also acknowledge comments by Brian Cheung on the manuscript.

\medskip

\small

\bibliographystyle{plain}
\bibliography{betagan}

\begin{thebibliography}{10}

\bibitem{arjovsky2017towards}
Martin Arjovsky and L{\'e}on Bottou.
\newblock Towards principled methods for training generative adversarial
  networks.
\newblock In {\em International Conference on Learning Representations}, 2017.

\bibitem{arjovsky2017wasserstein}
Martin Arjovsky, Soumith Chintala, and L{\'e}on Bottou.
\newblock Wasserstein {GAN}.
\newblock {\em arXiv preprint arXiv:1701.07875}, 2017.

\bibitem{bengio2009curriculum}
Yoshua Bengio, J{\'e}r{\^o}me Louradour, Ronan Collobert, and Jason Weston.
\newblock Curriculum learning.
\newblock In {\em Proceedings of the 26th annual international conference on
  machine learning}, pages 41--48. ACM, 2009.

\bibitem{BerthelotSM17BEGAN}
David Berthelot, Tom Schumm, and Luke Metz.
\newblock {BEGAN:} boundary equilibrium generative adversarial networks.
\newblock {\em CoRR}, abs/1703.10717, 2017.

\bibitem{denton2015deep}
Emily~L Denton, Soumith Chintala, Arthur Szlam, and Rob Fergus.
\newblock {Deep Generative Image Models using a Laplacian Pyramid of
  Adversarial Networks}.
\newblock In {\em Advances in neural information processing systems}, pages
  1486--1494, 2015.

\bibitem{goodfellow2014generative}
Ian Goodfellow, Jean Pouget-Abadie, Mehdi Mirza, Bing Xu, David Warde-Farley,
  Sherjil Ozair, Aaron Courville, and Yoshua Bengio.
\newblock Generative adversarial nets.
\newblock In {\em Advances in Neural Information Processing Systems}, pages
  2672--2680, 2014.

\bibitem{IoffeS15batchnormal}
Sergey Ioffe and Christian Szegedy.
\newblock Batch normalization: accelerating deep network training by reducing
  internal covariate shift.
\newblock {\em CoRR}, abs/1502.03167, 2015.

\bibitem{kaae2016amortised}
Casper Kaae~S{\o}nderby, Jose Caballero, Lucas Theis, Wenzhe Shi, and Ferenc
  Husz{\'a}r.
\newblock Amortised map inference for image super-resolution.
\newblock {\em arXiv preprint arXiv:1610.04490}, 2016.

\bibitem{lecun1998gradient}
Yann LeCun, L{\'e}on Bottou, Yoshua Bengio, and Patrick Haffner.
\newblock Gradient-based learning applied to document recognition.
\newblock {\em Proceedings of the IEEE}, 86(11):2278--2324, 1998.

\bibitem{liu2015faceattributes}
Ziwei Liu, Ping Luo, Xiaogang Wang, and Xiaoou Tang.
\newblock Deep learning face attributes in the wild.
\newblock In {\em Proceedings of International Conference on Computer Vision
  (ICCV)}, 2015.

\bibitem{narayanan2010sample}
Hariharan Narayanan and Sanjoy Mitter.
\newblock Sample complexity of testing the manifold hypothesis.
\newblock In {\em Advances in Neural Information Processing Systems}, pages
  1786--1794, 2010.

\bibitem{nowozin2016f}
Sebastian Nowozin, Botond Cseke, and Ryota Tomioka.
\newblock \emph{f}-{GAN}: Training generative neural samplers using variational
  divergence minimization.
\newblock In {\em Advances in Neural Information Processing Systems}, pages
  271--279, 2016.

\bibitem{radford2015unsupervised}
Alec Radford, Luke Metz, and Soumith Chintala.
\newblock Unsupervised representation learning with deep convolutional
  generative adversarial networks.
\newblock {\em arXiv preprint arXiv:1511.06434}, 2015.

\bibitem{salimans2016improved}
Tim Salimans, Ian Goodfellow, Wojciech Zaremba, Vicki Cheung, Alec Radford, and
  Xi~Chen.
\newblock Improved techniques for training {GAN}s.
\newblock In {\em Advances in Neural Information Processing Systems}, pages
  2226--2234, 2016.

\bibitem{saremi2013hierarchical}
Saeed Saremi and Terrence~J Sejnowski.
\newblock Hierarchical model of natural images and the origin of scale
  invariance.
\newblock {\em Proceedings of the National Academy of Sciences},
  110(8):3071--3076, 2013.

\bibitem{saremi2016correlated}
Saeed Saremi and Terrence~J Sejnowski.
\newblock Correlated percolation, fractal structures, and scale-invariant
  distribution of clusters in natural images.
\newblock {\em IEEE transactions on pattern analysis and machine intelligence},
  38(5):1016--1020, 2016.

\bibitem{sohl2015deep}
Jascha Sohl-Dickstein, Eric~A Weiss, Niru Maheswaranathan, and Surya Ganguli.
\newblock Deep unsupervised learning using nonequilibrium thermodynamics.
\newblock {\em arXiv preprint arXiv:1503.03585}, 2015.

\bibitem{zhao2016energy}
Junbo Zhao, Michael Mathieu, and Yann LeCun.
\newblock Energy-based generative adversarial network.
\newblock {\em arXiv preprint arXiv:1609.03126}, 2016.

\end{thebibliography}

\end{document}